\documentclass[conference]{IEEEtran}
\IEEEoverridecommandlockouts
\usepackage{cite}
\usepackage{amsmath,amssymb,amsfonts}
\usepackage{algorithmic}
\usepackage{graphicx}
\usepackage{textcomp}
\usepackage{siunitx}
\usepackage{xcolor}
\usepackage{blindtext}
\usepackage{hyperref}
\usepackage{subfiles} 

\def\BibTeX{{\rm B\kern-.05em{\sc i\kern-.025em b}\kern-.08em
    T\kern-.1667em\lower.7ex\hbox{E}\kern-.125emX}}
\begin{document}

\title{Image Based Food Energy Estimation \\ With Depth Domain Adaptation\\
}

\author{Gautham Vinod \quad  Zeman Shao \quad Fengqing Zhu \\
\textit{{School of Electrical and Computer Engineering, Purdue University, West Lafayette, Indiana USA}
}}

\maketitle

\begin{abstract}
Assessment of dietary intake has primarily relied on self-report instruments, which are prone to measurement errors. Dietary assessment methods have increasingly incorporated technological advances particularly mobile, image based approaches to address some of these limitations and further automation. 
Mobile, image-based methods can reduce user burden and bias by automatically estimating dietary intake from eating occasion images that are captured by mobile devices.
In this paper, we propose an ``Energy Density Map" which is a pixel-to-pixel mapping from the RGB image to the energy density of the food. We then incorporate the ``Energy Density Map" with an associated depth map that is captured by a depth sensor to estimate the food energy. 
The proposed method is evaluated on the Nutrition5k dataset. Experimental results show improved results compared to baseline methods with an average error of 13.29 kCal and average percentage error of 13.57\% between the ground-truth and the estimated energy of the food. 
\end{abstract}


\section{Introduction}
The ubiquity of smartphones and their integration into every facet of life has made it possible to monitor one's health status such as vital signs, exercise and sleep statistics, etc.  
What and how much a person eats and drinks in a day is another key contributor to one's overall health, yet very difficult to assess \cite{Konif2021}. 
Many existing mobile applications rely on manual user input to capture information about the foods they consumed \cite{Ze2021}. This follows the traditional dietary assessment method called the 24 hour Dietary Recalls where the participant either communicates with a dietitian or use a web-based tool to report their food intake in the last 24-hour period \cite{subar2012} which adds user burden and measurement error. 
For example, study has shown that multiple 24-hour recalls are needed to accurately capture the dietary intake \cite{Yunsheng2009}.
The development of image-based dietary assessment methods, particularly those using mobile devices, have the potential to reduce user burden and improve reporting bias compared to traditional approaches \cite{boushey2017, Zhu2010}.


Image-based methods have achieved impressive results in many applications for dietary assessment such as food recognition, food segmentation and food portion estimation \cite{Lo2020, He2021, Yarlagadda2021, Mao2021, Fang2018}. In this paper, we focus on food energy estimation which is challenging even for humans including domain experts (\textit{e.g.}, dietitians) to accurately estimate the energy of foods in an image without known physical references \cite{Shao2021, Lee2012}. 
Food energy estimation is closely related to image-based food portion estimation, which can be classified into three categories \cite{Lo2020}:
\begin{enumerate}
    \item \textbf{Single view image methods} where only a single view image of the eating scene is used for food portion estimation.
    \item \textbf{Multi-view images methods} where multi-view images of the eating scene are captured to extract depth information which is then used for food portion estimation. Usually, the depth is estimated from multiple images of the same scene to recover some of the 3D information of the foods from the 2D images. 
    \item \textbf{Depth based methods} where a depth map is captured by a depth sensor and aligned with the RGB image to estimate food portion size jointly.
\end{enumerate}
Relying only on a single-view image to estimate the food portion is a challenging task since the 3D information is lost during the projection from the real world coordinates to the image coordinates. 
Multi-view images methods typically increases user burden as the participant needs to capture images of the eating scene from different angles. 
Furthermore, multi-view images require additional post-processing, such as camera calibration and feature matching after image capture.
Depth map captured by the depth sensor represents pixel-wise distance from the object surface to the camera, which can be used to recover the 3D information. However, depth sensor is not commonly available on mobile devices, making it difficult to deploy multi-view based methods on existing mobile devices. 



Previously, we proposed a single-view based food portion estimation method by introducing the concept of an ``Energy Distribution Map"~\cite{Fang2018} which is a pixel to pixel mapping of the RGB image of the food to a map of how the energy of the food is distributed. A conditional Generative Adversarial Network (cGAN)  \cite{Mirza2014} is used to learn this mapping from the RGB image to the ``Energy Distribution Map." 
In \cite{Shao2021}, we further adopted the concept of the ``Energy Distribution Map" and combined its features with the features extracted from the RGB image to improve the estimation of food energy. The performance is shown to have improved when using a combination of these features as compared to using the ``Energy Distribution Map" alone. The method is evaluated on a real-world dataset collected from a dietary study where the ground-truth energy values are provided by registered dietitians. However, there are only 96 eating occasion images in the dataset.
In this paper, we first perform segmentation for each RGB image, and also adopt the ``Energy Distribution Map" as the ``Energy Density Map" which maps each pixel in the RGB image to an energy value, to more accurately represent the energy density of the different foods in the image.
We replace the RGB features with features extracted from a depth map to jointly perform food energy estimation. 
Our method is evaluated on a large public food image dataset, Nutrition5K, which consists of 3,493 food images with aligned depth maps captured by a depth sensor \cite{Thames2021}.

The main contributions of our paper can be summarized as follows:
\begin{enumerate}
    \item Propose an ``Energy Density Map" which maps each pixel in the RGB image to an energy value.
    \item Incorporate depth information from a depth sensor with information from the ``Energy Density Map" for joint learning of food energy in the image.
    \item Evaluate the proposed method on the public Nutrition5k dataset \cite{Thames2021}, which shows improved performance compared to previous methods using only ``Energy Density Map" and the RGB image. 

\end{enumerate} 




\section{Related Works}
Work related to food portion estimation typically tries to reconstruct 3D data from a 2D image, which is challenging because most of the 3D information is lost during the 3D to 2D projection. Existing methods are either based on extracting geometric information from the image or using a deep learning based approach.
\vspace{-2pt}
\subsection{Geometric Based Methods}
These methods estimate the food portion/volume from the geometry of objects in the image such as objects with known physical dimensions. 
For example, the shapes of containers in the image along with a physical reference object in the image such as a colored checkerboard of known dimensions are used as reference to estimate the volume of foods in the image \cite{Fang2015}. 
In \cite{Aizawa2013}, the authors use classification and estimation of serving sizes based on certain parameters of the image such as color, number of circles in the image, etc. and then these features are fed into an AdaBoost classifier \cite{freund1996} to estimate the serving sizes of the food. Mobile 3D range is used in \cite{Shang2011} where a structured laser grid is projected onto the food and a smartphone is used to capture the resulting image. The laser grid lines are used to create a 3D depth map of the food.
Other methods try to estimate the 3D scene using multiple images \cite{Puri2009, KONG2012,Dehais2013}.
The food volume is estimated in \cite{Chen2012} based on a reference object which is a circular dining plate and a user defined 3D model which represents the food image. This 3D model is projected onto the 2D image. The model uses the circular plate and the chosen 3D model to estimate the food volume.

\subsection{Deep Learning Based Methods}
Most deep learning based methods use Neural Networks to estimate the food portion. In \cite{Yunus2019} a CNN network is used for image classification and another network is used for food attribute estimation using vector space embedding \cite{mikolov2013}. In \cite{Pouladzadeh2016}, a CNN is used for food classification and then the portion is estimated based on a physical reference object in the image or the mobile camera's sensor measurements. A top and side view image are used in \cite{Liang2017} along with the Faster R-CNN network \cite{Ren2015} to identify and localize the food and a calibration object from these images. Segmentation is performed next to calculate the food volume based on the calibration object in the image. The concept of energy value per unit mass was used in \cite{Thames2021} and \cite{Ma2022}, which treats portion estimation as a classification task since each food has its own energy per unit mass which is a property inherent to the specific food.




\section{Method}
\subsection{Overview of Our Method}

\begin{figure*}
\centering
\includegraphics[width=\textwidth]{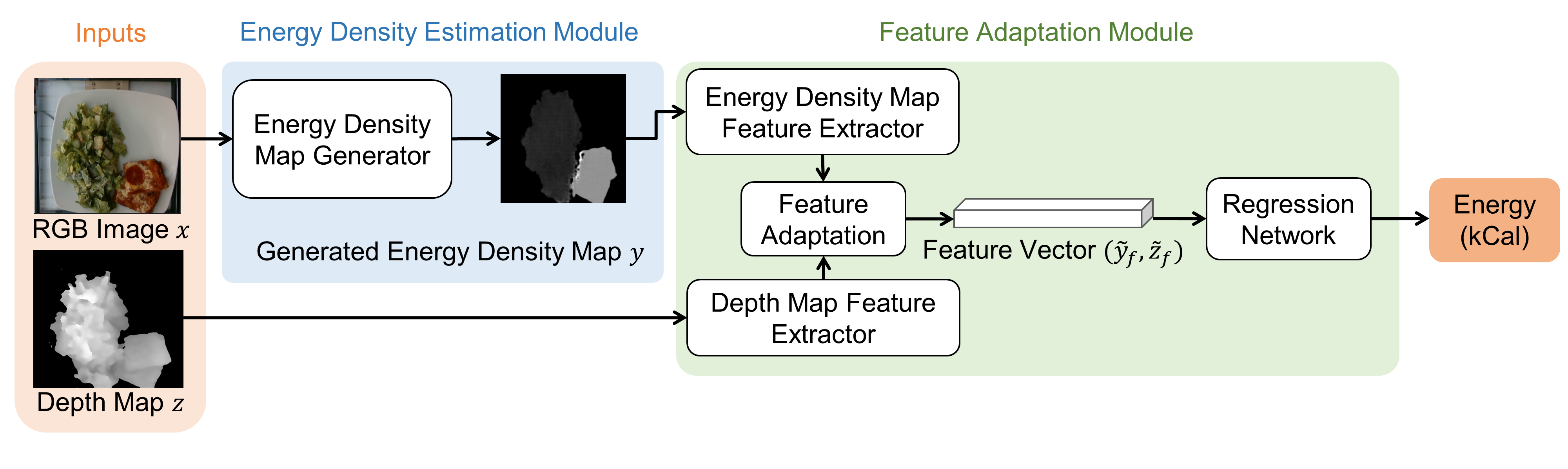}
\caption{\textbf{Overview of proposed method.} The Energy Density Estimation Module uses a generative model to estimate the Energy Density Map $y$ from the RGB input $x$. The Feature Adaptation Module extracts the features from the Energy Density Map $y$ and the Depth Map $z$, separately and combines them after normalizing the features. Finally, a regression network produces a single output value which is the estimated energy value of the food in the input image.}
\label{fig:overview}
\end{figure*}

Our method is based on the idea that energy of foods in an image can be estimated using the energy density, which is the energy per pixel in the image. We also incorporate the depth map which contains the physical distance information between the camera and the objects in the image. By keeping the distance between the camera and the objects (eating scene in this case) fixed, the depth map provides information about the relative sizes and shapes of the foods in the image.
Our method consists of 2 modules - an Energy Density Estimation Module, and a Feature Adaptation Module. The Energy Density Estimation Module consists of a generative model that maps the RGB image $x$ to the Energy Density Map $y$. The Feature Adaptation Module consists of 2 feature extracting networks which extract features from the generated energy density map $y$ and the depth map $z$ separately. The extracted features from the energy density map $y_f$ and the depth map $z_f$ are then normalized and concatenated. The normalized and concatenated features $(\tilde{y}_f, \tilde{z}_f)$ are then passed through a regression network to estimate the energy value of foods in the image. 
During training, the Energy Density Estimation Module is first trained to generate the Energy Density Map $y$. Once the generative model is trained, it estimates the Energy Density Map for all the images in our dataset which is then used as one of the inputs for the Feature Adaptation Module. The Feature Adaptation Module is trained to use the generated Energy Density Maps and the depth maps to estimate the food energy in the image. However, during the inference, an RGB image is first fed into the Energy Density Estimation Module and the output of this module along with the depth map are served as inputs for the Feature Adaptation Module, \textit{i.e.}, the inference is end-to-end.
Figure \ref{fig:overview} shows an overview of the proposed method.

\subsection{Energy Density Map}
The Energy Density Map $y$ is a 1-channel image where each pixel value is representative of the energy of the food at that pixel location in the image. During the energy estimation we use a conditional Generative Adversarial Netowrk (cGAN) \cite{Isola2017} to generate this Energy Density Map. In order to train this generative model we need to first obtain the ``ground-truth" energy density map.

To create the ground-truth energy density map for training, we first segment the food images so that we know the location of each individual food in the image. The segmentation step is necessary because the energy density of each food or ingredient can be different. For example, if we consider a small piece of meat and a small piece of spinach, both of the same size, the ground-truth energy value of the meat should be higher than that of the spinach. 
We use Google's Seefood mobile food segmenter \cite{Google} to perform the food segmentation, which adopts the DeepLab-V3 \cite{chen2018} network architecture with a MobileNet-v2 \cite{sandler2018} backbone.

Suppose a food image has $N$ foods or ingredients, then let $k$ denote the $k$-th food item in the image, $k \in [1,N]$. Let $e_k$ denote the energy value of the $k$-th food or ingredient in the image. Let pixel $x(i,j)$ correspond to a pixel in row $i$ and column $j$ of the RGB image $x$. The segmentation network gives us the pixel locations of each food $k$ in the image. Further, if $x(i,j)$ contains a food or ingredient $k$, then the corresponding pixel $y(i,j)$ of the energy density map has a value that is equal to 
\begin{equation*}
    y(i,j) = \frac{e_k}{\text{total number of pixels in food }k}
\end{equation*}
This process is repeated for each food or ingredient in the image. 
Once all the images in the dataset have a corresponding energy density map, these maps are scaled to the range of $[0,255]$ over the whole dataset. Therefore, this energy density map can be described as the distribution of a scaled value of the ground-truth food energy over the image.

We used the same training method as described in \cite{Shao2021} where the RGB image serves as the input to the generative model which outputs a generated Energy Density Map. 
  

\subsection{Joint Learning from Energy Density and Depth Map}

The energy density map $y$ provides information about how the energy of the foods are distributed over the area occupied by the foods in the image.
Previously, we extracted features from the Energy Density Map and the RGB image and then normalized and combined these features. The combination of these features are then used as inputs for a regression network to produce a final estimate of the food portion. In this paper, the features extracted from the depth map $z$ which contains information about the depth and shape of the food in the image, and hence we use these features instead of the features from the RGB image. We also evaluate the performance of the energy estimation when considering all three features, \textit{i.e.}, energy density, depth and RGB.

The depth map in the Nutrition5k Dataset is a 1 channel image with the same dimensions as the RGB image. Each pixel in the depth map is in the range of $[0, 65,535]$, which represents a scaled estimate of the distance between the camera and the object in that pixel.
The scale factor is $\SI{1e-4}{\meter}$, which means that if an object is $\SI{10}{\cm}$ away from the camera, then the corresponding pixels in the image have a value of $1,000$. We apply post-processing to the depth map from the depth sensor including dilation to smooth the foreground and morphological closing to fill any missing values in the depth map.
The energy density map and the depth map are then normalized to $[-1, 1]$ so that the features extracted from them can be concatenated.


\subsection{Food Energy Estimation}

The feature adaption module contains 2 feature extractors - The Energy Density Map feature extractor and the Depth Map feature extractor. Both use the VGG-16 \cite{Simonyan2014} network as their backbone. Previously, VGG-16 was used as the backbone to extract features from the Energy Density Map \cite{Shao2021} and Resnet-50 was used as the RGB feature extractor \cite{He2016}. Since extracting features from a single channel depth map is easier than extracting features from an RGB image, VGG-16 is sufficient in this case. The final fully-connected layers of the VGG-16 networks are removed. In each case, for the depth map and for the energy density map, the output of each VGG-16 feature extractor is a $7 \times 7 \times 512$ feature tensor.   

The feature tensors extracted from the Depth map and the Energy Density map are concatenated in the feature adaptation module. Since the features are from different domains, they are normalized before concatenation. We tried different methods of normalization such as Z-score normalization, Layer Normalization \cite{Ba2016} (layer normalization helps in normalizing the distributions of intermediate layers) in the regression network and a combination of Layer Normalization and Group Normalization (group normalization organizes channels into different groups and normalizes the distribution among them) \cite{Wu2018}.
Finally, these normalized features are passed through the same network used in \cite{Shao2021} which consists of 2 fully connected layers with normalization layers in between, which outputs a single value of the estimated energy of the food. 




\section{Experimental Results}
\subsection{Dataset Curation}

\begin{figure}
    \centering
    \includegraphics[width=\linewidth]{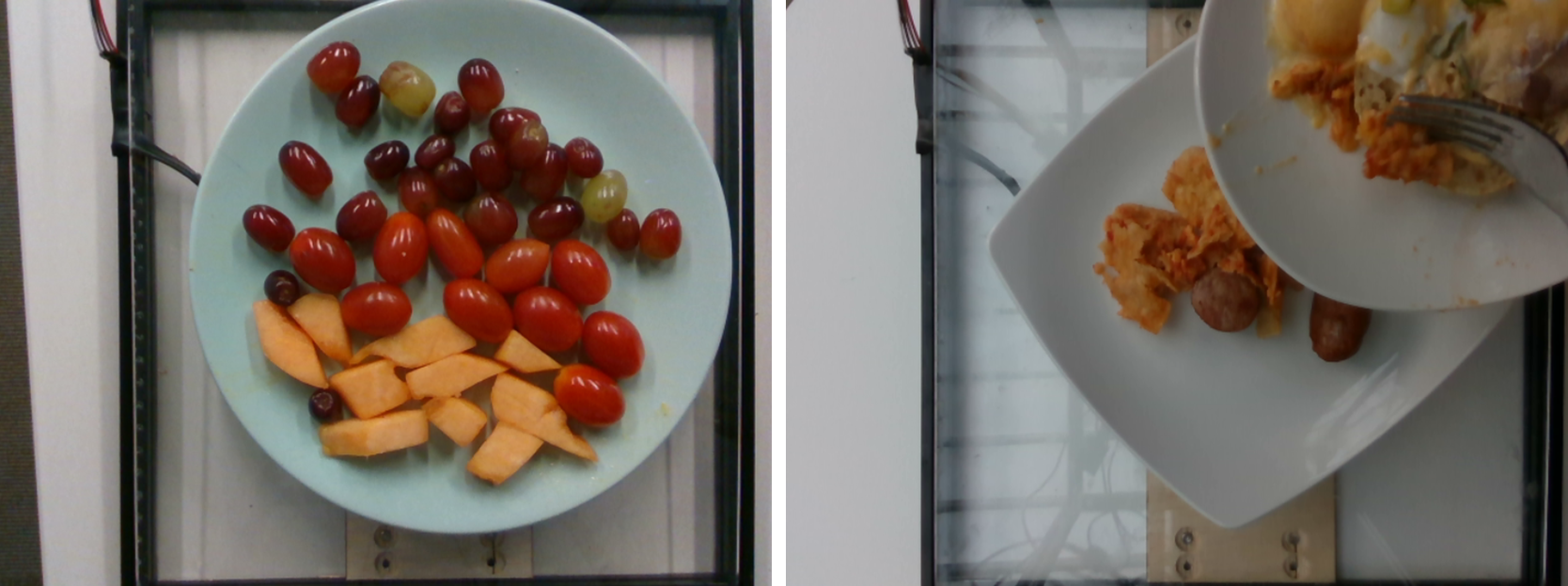}
    \caption{\textbf{Erroneous ground-truth data.} (Left) The ground-truth ingredients only contain Cantaloupes and Tomatoes but the image also contains grapes. (Right) Overlapping dishes where the food is not clearly visible because of improper image capture.}
    \label{fig:wrong_ground_truth}
\end{figure}

The Nutrition5k Dataset has 3,493 images that have overhead depth map data. In these 3,493 images, there are many dishes that have a lot of individual ingredients including many that are hard for a segmentation network to detect such as ketchup, mayonnaise, salt and pepper, oil, etc. As we mentioned in the previous section, the generation of the energy density map requires segmentation of the image. One of the limitations of the See-food segmentation network \cite{Google} is that the output classes are vague categories such as ``Leafy Greens", ``Starchy Vegetables", etc. making it difficult to group the ingredients in the dataset into these categories. Therefore, to address these challenges we choose a subset of the Nutrition5k Dataset where the segmentation network can achieve reasonable results. This subset of images consists of images with less than 3 food ingredients. In addition, we also removed images where one ingredient is completely covered by another ingredient and ingredients that do not have consistent appearance in all images such as tofu. 

\begin{figure}
    \centering
    \includegraphics[width=\linewidth]{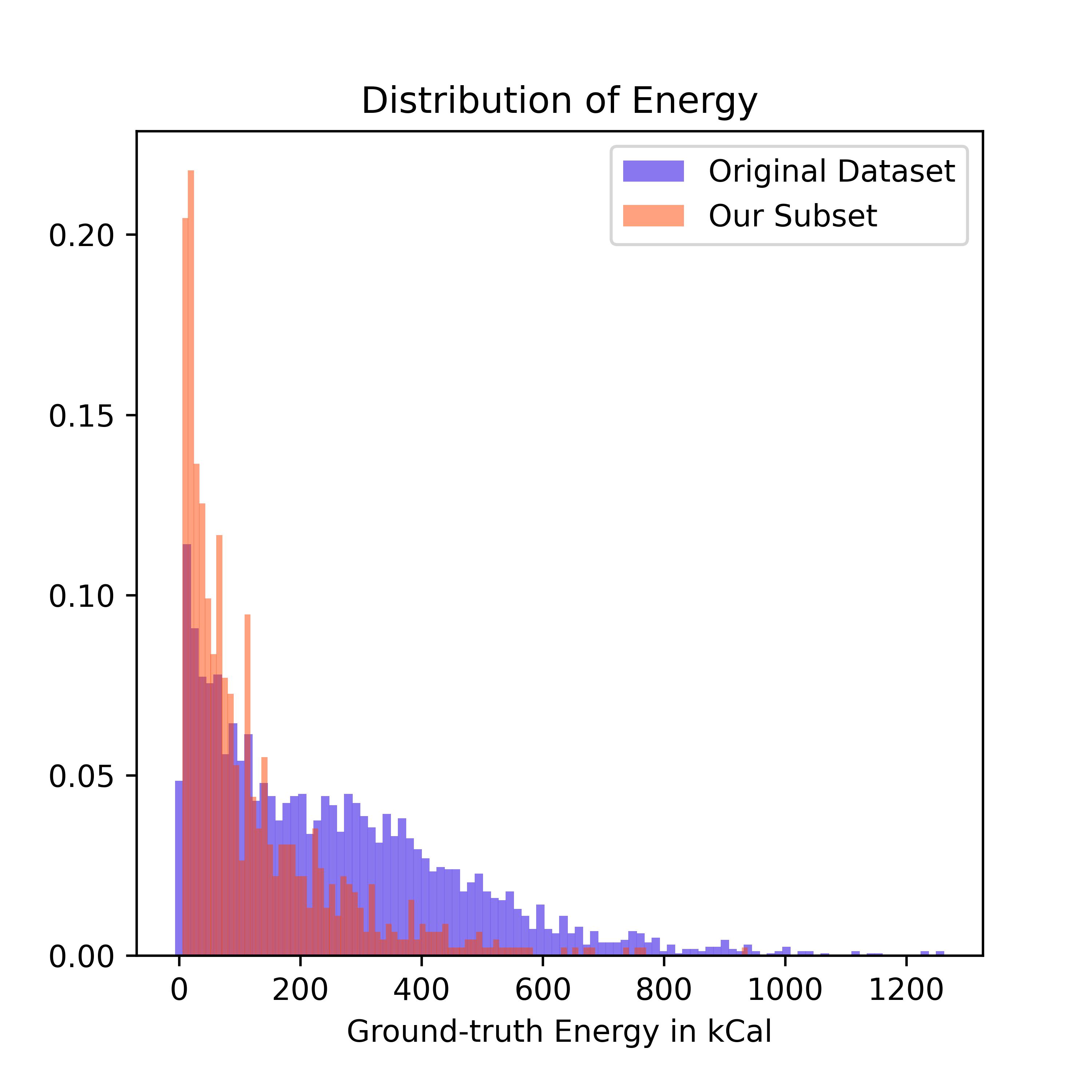}
    \caption{\textbf{Comparison of the distribution of energy in the datasets.} The graph shows that our subset has a higher concentration of images in the low ground-truth energy region as compared to the original dataset. The distribution also shows that the range of energy in our subset covers majority of the energy values in the original dataset.}
    \label{fig:distribution_comparision}
    \vspace{-4mm}
\end{figure}
There are a few instances where the ground-truth data provided in the dataset is incorrect such as the images in Figure \ref{fig:wrong_ground_truth}, which are excluded from our subset. We also removed images where the ground-truth energy value of the dish is below 10 kCal since there are only few images in the subset that are in this range of $0-10$ kCal. It would be challenge for the network to learn from limited data in this energy range.  
In the end, our subset contains 909 images. 
Figure \ref{fig:distribution_comparision} shows the distribution of the ground-truth energy values of the dataset (images with depth map available) and  our subset. 
The energy values in our subset approximately range from 10 kCal to 1,000 kCal while the range of the energy in the original dataset excluding the outliers (1.4\% of the data which is more than 3 standard deviations away from the mean) is around $[0, 800]$ kCal. Excluding the outliers in the original dataset, the range of our subset covers the majority of energy values in the original dataset. 
Since our subset has a good representation of energy range in the original dataset, it is suitable for evaluating the performance of the proposed method. 


\subsection{Experimental Results on the Nutrition5k Dataset}

The feature extractor for both the energy density map and the depth map use the VGG-16 network that is pre-trained on the ImageNet dataset \cite{Deng2009}. The feature extractor and the regression network are trained together, \textit{i.e.}, the entire feature adaptation module in Figure \ref{fig:overview} is trained together. The network is supervised on the sum of the L1 loss and the Mean Squared Error loss between the estimated and ground-truth energy of the food.
The Adam optimizer \cite{Kingma2014} is used with an initial learning rate of $5e^{-5}$. 
The output of the network is a scaled version of the estimated energy. We used 300 as the scaling factor for the output of the network, therefore the output of the network is the estimated energy scaled by our scaling factor. This output is scaled back to represent the estimated energy value. The scaling factor was determined experimentally.


The training and testing split is chosen by stratifying the 909 images so that the training data and the testing data have a similar distribution. An 80:20 stratified split is used resulting a total of 731 training images and 175 testing images. The learning rate is reduced by a factor of 0.8 for every 10 epochs and all models were trained for approximately 50 epochs each. We used the Mean Absolute Error (MAE) and Mean Absolute Percentage Error (MAPE) to measure the performance which are defined as
\begin{align*}
    \text{MAE} = \frac{1}{N}\sum_{i=1}^N|\hat{e} - e| \\
    \text{MAPE} = \frac{1}{N}\sum_{i=1}^N\frac{|\hat{e} - e|}{e} \times 100
\end{align*}
where $\hat{e}$ is the estimated energy of all foods in the image, $e$ is the ground-truth energy value, and $N$ is the total number of images in the test set.
The results in Table \ref{tab:results} shows the performance of our proposed method. The first three row show the baseline results using the features extracted from the RGB image $x_f$, the energy density map $y_f$, and the depth map $z_f$, respectively. The next two rows show the the combined features from the RGB image $x_f$ and the energy density map $y_f$ with and without feature adaptation. We then show the results of combining the energy density map $y_f$ and the depth map $z_f$ with and without feature adaptation. Finally, we show the result for combined features from the RGB image, energy density map and the depth map.  


\renewcommand{\arraystretch}{1.5}
\begin{table}
    \centering
    \caption{Comparison of the performance of features extracted from the RGB image $x_f$, energy density map $y_f$, and the depth map $z_f$ and their different combinations. $\tilde{x}_f$, $\tilde{y}_f$, and $\tilde{z}_f$ represents the normalized features.}
    \begin{tabular}[width=\linewidth]{|c c c|}
        \hline
        \textbf{Method} & \textbf{MAE (kCal)} & \textbf{MAPE (\%)} \\
        \hline
         $x_f$ & 26.85 & 40.64 \\
         $y_f$ & 13.35 & 16.90 \\
         $z_f$ & 76.86 & 133.88 \\
         \hline
         $(x_f, y_f)$ & 17.54 & 23.20 \\
         $(\tilde{x}_f, \tilde{y}_f)$ & 14.65 & 16.88 \\
         \hline
         $(y_f, z_f)$ & 15.83 & 27.04\\
         $(\tilde{y}_f, \tilde{z}_f)$ & 13.29 & 13.57\\
         \hline
         $(\tilde{x}_f, \tilde{y}_f, \tilde{z}_f)$ & 12.75 & 16.83 \\
         \hline
    \end{tabular}
    \label{tab:results}
\end{table}


From the results in Table \ref{tab:results}, we observed that:
\begin{enumerate}
    \item From the first three rows, it is clear that features from the Energy Density Map contributes the most to reducing the energy estimation error.
    \item The depth map by itself does not have sufficient information to accurately estimate the energy of the food. Since the depth and the energy density maps are in different domains, combining them without normalization degrades the performance compared to using the Energy Density Map alone.
    \item The best performance is obtained when the features from the depth map and the energy density map are combined and normalized, resulting in a MAPE of 13.57\%. When the RGB, depth and, energy density map features are combined, the MAE is lower than when only the depth map and the Energy Density Map are combined. However, MAPE is a better indicator of performance in this subset since the low ground-truth energy values of the food push the MAPE to be higher even for a small difference in MAE. 
\end{enumerate}

\subsection{Comparison to Nutrition5k Results}
Three portion estimation methods are reported using the Nutrition5k dataset in \cite{Thames2021}. 
\begin{enumerate}
    \item 2D Direct Prediction - The RGB image is used as an input to a regression network with the Inception v3 \cite{Szegedy2015} network as its backbone.
    
    Reported MAE (kCal) / MAPE: 70.6 / 26.1\%
    \item Depth as the 4-th channel - The depth map is added as another channel to the RGB input and then this RGB-D image is sampled to form a 3-channel input to the regression network. 
    
    Reported MAE (kCal) / MAPE: 47.6 / 18.8\%
    \item Volume Scalar - The mass of the food is approximated using certain physical approximations and the depth map, and then multiplied with a network that predicts the calories per gram of the food. 
    
    Reported MAE (kCal) / MAPE: 41.3 / 16.5\%
\end{enumerate}

The mean energy value (124.96 kCal) in our subset is  lower than the full dataset (254.94 kCal) and hence we would expect our MAPE to be higher in our subset. This is because an estimate of 20 kCal for a food that has a ground-truth energy of 10 kCal will produce an MAPE of 100\% but an MAE of only 10 kCal. However, we see that our proposed method combining the Energy Density Map and the depth map has a lower MAPE of 13.57\% than the reported results of 18.8\% when depth map is used in \cite{Thames2021} . 

\section{Conclusion}
We proposed a method that combines the information from an Energy Density Map and a depth map to estimate energy for the foods in an image, where the Energy Density Map is generated from an RGB input. Our preliminary experiments showed promising results on a subset of the Nutrition5K dataset. Our future work will focus on evaluating our method on the complete Nutrition5K dataset and food images captured in dietary studies.



\bibliographystyle{IEEEtran} 
\bibliography{references} 

\end{document}